# Continuous Inverse Optimal Control with Locally Optimal Examples

**Sergey Levine**                                    SVLEVINE@STANFORD.EDU
**Vladlen Koltun**                                   VLADLEN@STANFORD.EDU
Computer Science Department, Stanford University, Stanford, CA 94305 USA

## Abstract

Inverse optimal control, also known as inverse reinforcement learning, is the problem of recovering an unknown reward function in a Markov decision process from expert demonstrations of the optimal policy. We introduce a probabilistic inverse optimal control algorithm that scales gracefully with task dimensionality, and is suitable for large, continuous domains where even computing a full policy is impractical. By using a local approximation of the reward function, our method can also drop the assumption that the demonstrations are globally optimal, requiring only local optimality. This allows it to learn from examples that are unsuitable for prior methods.

## 1. Introduction

Algorithms for inverse optimal control (IOC), also known as inverse reinforcement learning (IRL), recover an unknown reward function in a Markov decision process (MDP) from expert demonstrations of the corresponding policy. This reward function can be used to perform apprenticeship learning, generalize the expert's behavior to new situations, or infer the expert's goals (Ng & Russell, 2000). Performing IOC in continuous, high-dimensional domains is challenging, because IOC algorithms are usually much more computationally demanding than the corresponding "forward" control methods. In this paper, we present an IOC algorithm that efficiently handles deterministic MDPs with large, continuous state and action spaces by considering only the shape of the learned reward function in the neighborhood of the expert's demonstrations.

Since our method only considers the shape of the reward function around the expert's examples, it does not integrate global information about the reward along alternative paths. This is analogous to trajectory optimization methods, which solve the forward control problem by finding a local optimum. However, while the lack of global optimality is a disadvantage for solving the forward problem, it can actually be advantageous in IOC. This is because it removes the assumption that the expert demonstrations are globally optimal, thus allowing our algorithm to use examples that only exhibit local optimality. For complex tasks, human experts might find it easier to provide such locally optimal examples. For instance, a skilled driver might execute every turn perfectly, but still take a globally suboptimal route to the destination.

Our algorithm optimizes the approximate likelihood of the expert trajectories under a parameterized reward. The approximation assumes that the expert's trajectory lies near a peak of this likelihood, and the resulting optimization finds a reward function under which this peak is most prominent. Since this approach only considers the shape of the reward around the examples, it does not require the examples to be globally optimal, and remains efficient even in high dimensions. We present two variants of our algorithm that learn the reward either as a linear combination of the provided features, as is common in prior work, or as a nonlinear function of the features, as in a number of recent methods (Ratliff et al., 2009; Levine et al., 2010; 2011).

## 2. Related Work

Most prior IOC methods solve the entire forward control problem in the inner loop of an iterative procedure (Abbeel & Ng, 2004; Ratliff et al., 2006; Ziebart, 2010). Such methods often use an arbitrary, possibly approximate forward solver, but this solver must be used numerous times during the learning process, making reward learning significantly more costly than the forward problem. Dvijotham and Todorov avoid repeated calls to a forward solver by directly learning a value function (Dvijotham & Todorov, 2010). However, this requires value function bases to impose structure on the solution, instead of the more common reward bases. Good value function bases are difficult to construct and are not portable across domains. By only considering the reward around the expert's trajectories, our method removes the need to repeatedly solve a difficult forward problem, without losing the ability to utilize informative reward features.

More efficient IOC algorithms have been proposed for the special case of linear dynamics and quadratic rewards (LQR) (Boyd et al., 1994; Ziebart, 2010). However, unlike in the forward case, LQR approaches are difficult to generalize to arbitrary inverse problems, because learning a quadratic reward matrix around an example path does not readily generalize to other states in a non-LQR task. Because of this, such methods have only been applied to tasks that conform to the LQR assumptions (Ziebart, 2010). Our method also uses a quadratic expansion of the reward function, but instead of learning the values of a quadratic reward matrix directly, it learns a general parameterized reward using its Hessian and gradient. As we show in Section 5, a particularly efficient variant of our algorithm can be derived when the dynamics are linearized, and this derivation can in fact follow from standard LQR assumptions. However, this approximation is not required, and the general form of our algorithm does not assume linearized dynamics.

Most previous methods also assume that the expert demonstrations are globally optimal or near-optimal. Although this makes the examples more informative insofar as the learning algorithm can extract relevant global information, it also makes such methods unsuitable for learning from examples that are only locally optimal. As shown in the evaluation, our method can learn rewards even from locally optimal examples.

## 3. Background

We address deterministic, fixed-horizon control tasks with continuous states $\mathbf{x} = (\mathbf{x}_1, \ldots, \mathbf{x}_T)^{\mathrm{T}}$, continuous actions $\mathbf{u} = (\mathbf{u}_1, \ldots, \mathbf{u}_T)^{\mathrm{T}}$, and discrete time. Such tasks are characterized by a dynamics function $\mathcal{F}$, which we define as

$$\mathcal{F}(\mathbf{x}_{t-1}, \mathbf{u}_t) = \mathbf{x}_t,$$

as well as a reward function $r(\mathbf{x}_t, \mathbf{u}_t)$. Given the initial state $\mathbf{x}_0$, the optimal actions are given by

$$\mathbf{u} = \arg \max_{\mathbf{u}} \sum_t r(\mathbf{x}_t, \mathbf{u}_t).$$

IOC aims to find a reward function $r$ under which the optimal actions match the expert's demonstrations,

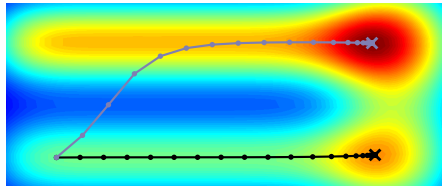

Figure 1. A trajectory that is locally optimal but globally suboptimal (black) and a globally optimal trajectory (grey). While prior methods usually require globally optimal examples, our approach can use examples that are only locally optimal. Warmer colors indicate higher reward.

given by $\mathcal{D} = \{(\mathbf{x}_0^{(1)}, \mathbf{u}^{(1)}), \ldots, (\mathbf{x}_0^{(n)}, \mathbf{u}^{(n)})\}$. The algorithm might also be presented with reward features $\mathbf{f} : (\mathbf{x}_t, \mathbf{u}_t) \rightarrow \mathbb{R}$ that can be used to represent the unknown reward $r$. Unfortunately, real demonstrations are rarely perfectly optimal, so we require a model for the expert's behavior that can explain suboptimality or "noise." We employ the maximum entropy IRL (MaxEnt) model (Ziebart et al., 2008), which is closely related to linearly-solvable MDPs (Dvijotham & Todorov, 2010). Under this model, the probability of the actions $\mathbf{u}$ is proportional to the exponential of the rewards encountered along the trajectory:[1]

$$P(\mathbf{u}|\mathbf{x}_0) = \frac{1}{Z} \exp \left( \sum_t r(\mathbf{x}_t, \mathbf{u}_t) \right), \qquad (1)$$

where $Z$ is the partition function. Under this model, the expert follows a stochastic policy that becomes more deterministic when the stakes are high, and more random when all choices have similar value. In prior work, the log likelihood derived from Equation 1 was maximized directly. However, computing the partition function $Z$ requires finding the complete policy under the current reward, using a variant of value iteration (Ziebart, 2010). In high dimensional spaces, this becomes intractable, since this computation scales exponentially with the dimensionality of the state space.

In the following sections, we present an approximation to Equation 1 that admits efficient learning in high dimensional, continuous domains. In addition to breaking the exponential dependence on dimensionality, this approximation removes the requirement that the example trajectories be globally optimal, and only requires approximate local optimality. An example of a locally optimal but globally suboptimal trajectory is shown in Figure 1: although another path has a higher total reward, any local perturbation of the trajectory decreases the reward total.

---

[1] In the case of multiple example trajectories, their probabilities are simply multiplied together to obtain $P(\mathcal{D})$.

## 4. IOC with Locally Optimal Examples

To evaluate Equation 1 without computing the partition function $Z$, we apply the Laplace approximation, which locally models the distribution as a Gaussian (Tierney & Kadane, 1986). Note that this is not equivalent to modeling the reward function itself as a Gaussian, since Equation 1 uses the sum of the rewards along a path. In the context of IOC, this corresponds to assuming that the expert performs a *local* optimization when choosing the actions $\mathbf{u}$, rather than global planning. This assumption is strictly less restrictive than the assumption of global optimality.

Using $r(\mathbf{u})$ to denote the sum of rewards along path $(\mathbf{x}_0, \mathbf{u})$, we can write Equation 1 as

$$P(\mathbf{u}|\mathbf{x}_0) = e^{r(\mathbf{u})} \left[ \int e^{r(\tilde{\mathbf{u}})} d\tilde{\mathbf{u}} \right]^{-1}.$$

We approximate this probability using a second order Taylor expansion of $r$ around $\mathbf{u}$:

$$r(\tilde{\mathbf{u}}) \approx r(\mathbf{u}) + (\tilde{\mathbf{u}} - \mathbf{u})^{\mathrm{T}} \frac{\partial r}{\partial \mathbf{u}} + \frac{1}{2}(\tilde{\mathbf{u}} - \mathbf{u})^{\mathrm{T}} \frac{\partial^2 r}{\partial \mathbf{u}^2}(\tilde{\mathbf{u}} - \mathbf{u}).$$

Denoting the gradient $\frac{\partial r}{\partial \mathbf{u}}$ as $\mathbf{g}$ and the Hessian $\frac{\partial^2 r}{\partial \mathbf{u}^2}$ as $\mathbf{H}$, the approximation to Equation 1 is given by

$$P(\mathbf{u}|\mathbf{x}_0) \approx e^{r(\mathbf{u})} \left[ \int e^{r(\mathbf{u}) + (\tilde{\mathbf{u}} - \mathbf{u})^{\mathrm{T}} \mathbf{g} + \frac{1}{2}(\tilde{\mathbf{u}} - \mathbf{u})^{\mathrm{T}} \mathbf{H}(\tilde{\mathbf{u}} - \mathbf{u})} d\tilde{\mathbf{u}} \right]^{-1}$$

$$= \left[ \int e^{-\frac{1}{2}\mathbf{g}^{\mathrm{T}} \mathbf{H}^{-1} \mathbf{g} + \frac{1}{2}(\mathbf{H}(\tilde{\mathbf{u}} - \mathbf{u}) + \mathbf{g})^{\mathrm{T}} \mathbf{H}^{-1}(\mathbf{H}(\tilde{\mathbf{u}} - \mathbf{u}) + \mathbf{g})} d\tilde{\mathbf{u}} \right]^{-1}$$

$$= e^{\frac{1}{2}\mathbf{g}^{\mathrm{T}} \mathbf{H}^{-1} \mathbf{g}} |-\mathbf{H}|^{\frac{1}{2}} (2\pi)^{-\frac{d_\mathbf{u}}{2}},$$

from which we obtain the approximate log likelihood

$$\mathcal{L} = \frac{1}{2}\mathbf{g}^{\mathrm{T}} \mathbf{H}^{-1} \mathbf{g} + \frac{1}{2} \log |-\mathbf{H}| - \frac{d_\mathbf{u}}{2} \log 2\pi. \quad (2)$$

Intuitively, this likelihood indicates that reward functions under which the example paths have small gradients and large negative Hessians are more likely. The magnitude of the gradient corresponds to how close the example is to a local peak in the (total) reward landscape, while the Hessian describes how steep this peak is. For a given parameterization of the reward, we can learn the most likely parameters by maximizing Equation 2. In the next section, we discuss how this objective and its gradients can be computed efficiently.

## 5. Efficient Likelihood Optimization

We can optimize Equation 2 directly with any optimization method. The computation is dominated by the linear system $\mathbf{H}^{-1}\mathbf{g}$, so the cost is cubic in the path length $T$ and the action dimensionality. We will describe two approximate algorithms that evaluate the likelihood in time linear in $T$ by linearizing the dynamics. This greatly speeds up the method on longer examples, though it should be noted that modern linear solvers are well optimized for symmetric matrices such as $\mathbf{H}$, making it quite feasible to evaluate the likelihood without linearization for moderate length paths.

To derive the approximate linear-time solution to $\mathbf{H}^{-1}\mathbf{g}$, we first express $\mathbf{g}$ and $\mathbf{H}$ in terms of the derivatives of $r$ with respect to $\mathbf{x}$ and $\mathbf{u}$ individually:

$$\mathbf{g} = \underbrace{\frac{\partial r}{\partial \mathbf{u}}}_{\tilde{\mathbf{g}}} + \underbrace{\frac{\partial \mathbf{x}}{\partial \mathbf{u}}}_{\mathbf{J}}{}^{\mathrm{T}} \underbrace{\frac{\partial r}{\partial \mathbf{x}}}_{\hat{\mathbf{g}}},$$

$$\mathbf{H} = \underbrace{\frac{\partial^2 r}{\partial \mathbf{u}^2}}_{\tilde{\mathbf{H}}} + \underbrace{\frac{\partial \mathbf{x}}{\partial \mathbf{u}}}_{\mathbf{J}}{}^{\mathrm{T}} \underbrace{\frac{\partial^2 r}{\partial \mathbf{x}^2}}_{\hat{\mathbf{H}}} \underbrace{\frac{\partial \mathbf{x}}{\partial \mathbf{u}}}_{\mathbf{J}}{}^{\mathrm{T}} + \underbrace{\frac{\partial^2 \mathbf{x}}{\partial \mathbf{u}^2}}_{\hat{\mathbf{H}}} \underbrace{\frac{\partial r}{\partial \mathbf{x}}}_{\hat{\mathbf{g}}}.$$

Since $r(\mathbf{x}_t, \mathbf{u}_t)$ depends only on the state and action at time $t$, $\tilde{\mathbf{H}}$ and $\hat{\mathbf{H}}$ are block diagonal, with $T$ blocks. To build the Jacobian $\mathbf{J}$, we differentiate the dynamics:

$$\frac{\partial \mathcal{F}}{\partial \mathbf{u}_t}(\mathbf{x}_{t-1}, \mathbf{u}_t) = \frac{\partial \mathbf{x}_t}{\partial \mathbf{u}_t} = \mathbf{B}_t,$$

$$\frac{\partial \mathcal{F}}{\partial \mathbf{x}_{t-1}}(\mathbf{x}_{t-1}, \mathbf{u}_t) = \frac{\partial \mathbf{x}_t}{\partial \mathbf{x}_{t-1}} = \mathbf{A}_t.$$

Future actions do not influence past states, so $\mathbf{J}$ is block upper triangular. Using the Markov property, we can express the nonzero blocks recursively:

$$\mathbf{J}_{t_1, t_2} = \frac{\partial \mathbf{x}_{t_1}}{\partial \mathbf{u}_{t_2}}{}^{\mathrm{T}} = \begin{cases} \mathbf{B}_{t_1}^{\mathrm{T}}, & t_1 = t_2 \\ \mathbf{J}_{t_1, t_1 - 1} \mathbf{A}_{t_1}^{\mathrm{T}}, & t_1 > t_2 \end{cases}$$

We can now write $\mathbf{g}$ and $\mathbf{H}$ almost entirely in terms of matrices that are block diagonal or block triangular. Unfortunately, the final second order term $\hat{\mathbf{H}}$ does not exhibit such convenient structure. In particular, the Hessian of the last state $\mathbf{x}_T$ with respect to the actions $\mathbf{u}$ can be arbitrarily dense. We will therefore disregard this term. Since $\hat{\mathbf{H}}$ is zero only when the dynamics are linear, this corresponds to linearizing the dynamics.

### 5.1. Direct Likelihood Evaluation

We first describe an approach for directly evaluating the likelihood under the assumption that $\hat{\mathbf{H}}$ is zero. We first exploit the structure of $\mathbf{J}$ to evaluate $\mathbf{J}\hat{\mathbf{g}}$ in time linear in $T$, which is essential for computing $\mathbf{g}$. This requires a simple recursion from $t = T$ to 1:

$$[\mathbf{J}\hat{\mathbf{g}}]_t \leftarrow \mathbf{B}_t^{\mathrm{T}}(\hat{\mathbf{g}}_t + \mathbf{z}_\mathbf{g}^{(t)}) \quad \mathbf{z}_\mathbf{g}^{(t+1)} \leftarrow \mathbf{A}_t^{\mathrm{T}}(\hat{\mathbf{g}}_t + \mathbf{z}_\mathbf{g}^{(t)}), \quad (3)$$

where $\mathbf{z}_\mathbf{g}$ accumulates the product of $\hat{\mathbf{g}}$ with the off-diagonal elements of $\mathbf{J}$. The linear system $\mathbf{h} = \mathbf{H}^{-1}\mathbf{g}$ can be solved with a stylistically similar recursion. We first use the assumption that $\tilde{\mathbf{H}}$ is zero to factor $\mathbf{H}$:

$$\mathbf{H} = (\tilde{\mathbf{H}}\mathbf{P} + \mathbf{J}\hat{\mathbf{H}})\mathbf{J}^{\mathrm{T}},$$

where $\mathbf{P}\mathbf{J}^{\mathrm{T}} = \mathbf{I}$. The nonzero blocks of $\mathbf{P}$ are

$$\mathbf{P}_{t,t} = \mathbf{B}_t^{\dagger} \qquad \mathbf{P}_{t,t-1} = -\mathbf{B}_t^{\dagger}\mathbf{A}_t,$$

and $\mathbf{B}_t^{\dagger}$ is a pseudoinverse of the potentially nonsquare matrix $\mathbf{B}_t$. This linear system is solved in two passes: an upward pass to solve $(\tilde{\mathbf{H}}\mathbf{P} + \mathbf{J}\hat{\mathbf{H}})\bar{\mathbf{h}} = \mathbf{g}$, and a downward pass to solve $\mathbf{J}^{\mathrm{T}}\mathbf{h} = \bar{\mathbf{h}}$. Each pass is a block generalization of forward or back substitution and, like the recursion in Equation 3, can exploit the structure of $\mathbf{J}$ to run in time linear in $T$. However, the upward pass must also handle the off-diagonal entries of $\mathbf{P}$ and the potentially nonsquare blocks of $\mathbf{J}$, which are not invertible. We therefore only construct a partial solution on the upward pass, with each $\bar{\mathbf{h}}_t$ expressed in terms of $\bar{\mathbf{h}}_{t-1}$. The final values are reconstructed on the downward pass, together with the solution $\mathbf{h}$. The complete algorithm computes $\mathbf{h} = \mathbf{H}^{-1}\mathbf{g}$ and the determinant $|-\mathbf{H}|$ in time linear in $T$, and is included in Appendix A of the supplement.

For a given parameterization of the reward, we determine the most likely parameters by maximizing the likelihood with gradient-based optimization (LBFGS in our implementation). This requires the gradient of Equation 2. For a reward parameter $\theta$, the gradient is

$$\frac{\partial \mathcal{L}}{\partial \theta} = \mathbf{h}^{\mathrm{T}}\frac{\partial \mathbf{g}}{\partial \theta} - \frac{1}{2}\mathbf{h}^{\mathrm{T}}\frac{\partial \mathbf{H}}{\partial \theta}\mathbf{h} + \frac{1}{2}\mathrm{tr}\left(\mathbf{H}^{-1}\frac{\partial \mathbf{H}}{\partial \theta}\right).$$

As before, the gradients of $\mathbf{g}$ and $\mathbf{H}$ can be expressed in terms of the derivatives of $r$ at each time step, which allows us to rewrite the gradient as

$$\frac{\partial \mathcal{L}}{\partial \theta} = \sum_{ti}\frac{\partial \tilde{\mathbf{g}}_{ti}}{\partial \theta}\mathbf{h}_{ti} + \sum_{ti}\frac{\partial \hat{\mathbf{g}}_{ti}}{\partial \theta}[\mathbf{J}^{\mathrm{T}}\mathbf{h}]_{ti} + \tag{4}$$

$$\frac{1}{2}\sum_{tij}\frac{\partial \tilde{\mathbf{H}}_{tij}}{\partial \theta}\left([\mathbf{H}^{-1}]_{ttij} - \mathbf{h}_{ti}\mathbf{h}_{tj}\right) +$$

$$\frac{1}{2}\sum_{tij}\frac{\partial \hat{\mathbf{H}}_{tij}}{\partial \theta}\left([\mathbf{J}^{\mathrm{T}}\mathbf{H}^{-1}\mathbf{J}]_{ttij} - [\mathbf{J}^{\mathrm{T}}\mathbf{h}]_{ti}[\mathbf{J}^{\mathrm{T}}\mathbf{h}]_{tj}\right) +$$

$$\frac{1}{2}\sum_{ti}\frac{\partial \tilde{\mathbf{g}}_{ti}}{\partial \theta}\sum_{t_1 t_2 jk}\left([\mathbf{H}^{-1}]_{t_1 t_2 jk} - \mathbf{h}_{t_1 j}\mathbf{h}_{t_2 k}\right)\breve{\mathbf{H}}_{t_1 t_2 jkti},$$

where $[\mathbf{H}^{-1}]_{ttij}$ denotes the $ij^{\mathrm{th}}$ entry in the block $tt$. The last sum vanishes if $\breve{\mathbf{H}}$ is zero, so all quantities can be computed in time linear in $T$. The diagonal blocks of $\mathbf{H}^{-1}$ and $\mathbf{J}^{\mathrm{T}}\mathbf{H}^{-1}\mathbf{J}$ can be computed while solving for $\mathbf{h} = \mathbf{H}^{-1}\mathbf{g}$, as shown in Appendix A of the supplement. To find the gradients for any reward parameterization, we can compute the gradients of $\hat{\mathbf{g}}$, $\tilde{\mathbf{g}}$, $\hat{\mathbf{H}}$, and $\tilde{\mathbf{H}}$, and then apply the above equation.

## 5.2. LQR-Based Likelihood Evaluation

While the approximate likelihood in Equation 2 makes no assumptions about the dynamics of the MDP, the algorithm in Section 5.1 linearizes the dynamics around the examples. This matches the assumptions of the commonly studied linear-quadratic regulator (LQR) setting, and suggests an alternative derivation of the algorithm as IOC in a linear-quadratic system, with linearized dynamics given by $\mathbf{A}_t$ and $\mathbf{B}_t$, quadratic reward matrices given by the diagonal blocks of $\hat{\mathbf{H}}$ and $\tilde{\mathbf{H}}$, and linear reward vectors given by $\hat{\mathbf{g}}$ and $\tilde{\mathbf{g}}$. A complete derivation of the resulting algorithm is presented in Appendix B of the supplement, and is similar to the MaxEnt LQR IOC algorithm described by Ziebart (Ziebart, 2010), with an addition recursion to compute the derivatives of the parameterized reward Hessians. Since the gradients are computed recursively, this method lacks the convenient form provided by Equation 4, but may be easier to implement.

## 6. Algorithms for IOC with Locally Optimal Examples

We can use the objective in Equation 2 to learn reward functions with a variety of representations. We present one variant that learns the reward as a linear combination of features, and a second variant that uses a Gaussian process to learn nonlinear reward functions.

### 6.1. Learning Linear Reward Functions

In the linear variant, the algorithm is provided with features $\mathbf{f}$ that depend on the state $\mathbf{x}_t$ and action $\mathbf{u}_t$. The reward is given by $r(\mathbf{x}_t, \mathbf{u}_t) = \theta^{\mathrm{T}}\mathbf{f}(\mathbf{x}_t, \mathbf{u}_t)$, and the weights $\theta$ are learned. Letting $\tilde{\mathbf{g}}^{(k)}$, $\hat{\mathbf{g}}^{(k)}$, $\tilde{\mathbf{H}}^{(k)}$, and $\hat{\mathbf{H}}^{(k)}$ denote the gradients and Hessians of each feature with respect to actions and states, the full gradients and Hessians are sums of these quantities, weighted by $\theta_k$ – e.g., $\tilde{\mathbf{g}} = \sum_k \tilde{\mathbf{g}}^{(k)}\theta_k$. The gradient of $\tilde{\mathbf{g}}$ with respect to $\theta_k$ is simply $\tilde{\mathbf{g}}^{(k)}$, and the gradients of the other matrices are given analogously. The likelihood gradient is then obtained from Equation 4.

When evaluating Equation 2, the log determinant of the negative Hessian is undefined when the determinant is not positive. This corresponds to the example path lying in a valley rather than on a peak of the energy landscape. A high-probability reward function will avoid such cases, but it is nontrivial to find an initial point for which the objective can be evaluated. We therefore add a dummy regularizer feature that ensures that the negative Hessian has a positive determinant. This feature has a gradient that is uniformly zero, and a Hessian equal to the negative identity.

The initial weight $\theta_r$ on this feature must be set such that the negative Hessians of all example paths are positive definite. We can find a suitable weight simply by doubling $\theta_r$ until this requirement is met. During the optimization, we must drive $\theta_r$ to zero in order to solve the original problem. In this way, $\theta_r$ has the role of a relaxation, allowing the algorithm to explore the parameter space without requiring the Hessian to always be negative definite. Unfortunately, driving $\theta_r$ to zero too quickly can create numerical instability, as the Hessians become ill-conditioned or singular. Rather than simply penalizing the regularizing weight, we found that we can maintain numerical stability and still obtain a solution with $\theta_r = 0$ by using the Augmented Lagrangian method (Birgin & Martínez, 2009). This method solves a sequence of maximization problems that are augmented by a penalty term of the form

$$\phi^{(j)}(\theta_r) = -\frac{1}{2}\mu^{(j)}\theta_r^2 + \lambda^{(j)}\theta_r,$$

where $\mu^{(j)}$ is a penalty weight, and $\lambda^{(j)}$ is an estimate of the Lagrange multiplier for the constraint $\theta_r = 0$. After each optimization, $\mu^{(j+1)}$ is increased by a factor of 10 if $\theta_r$ has not decreased, and $\lambda^{(j+1)}$ is set to

$$\lambda^{(j+1)} \leftarrow \lambda^{(j)} - \mu^{(j)}\theta_r.$$

This approach allows $\theta_r$ to decrease gradually with each optimization without using large penalty terms.

## 6.2. Learning Nonlinear Reward Functions

In the nonlinear variant of our algorithm, we represent the reward function as a Gaussian process (GP) that maps from feature values to rewards, as proposed by Levine et al. (Levine et al., 2011). The inputs of the Gaussian process are a set of inducing feature points $\mathbf{F} = [\mathbf{f}^1 \dots \mathbf{f}^n]^{\mathrm{T}}$, and the noiseless outputs $\mathbf{y}$ at these points are learned. The location of the inducing points can be chosen in a variety of ways, but we follow Levine et al. and choose the points that lie on the example paths, which concentrates the learning on the regions where the examples are most informative. In addition to the outputs $\mathbf{y}$, we also learn the hyperparameters $\lambda$ and $\beta$ that describe the GP kernel function, given by

$$k(\mathbf{f}^i, \mathbf{f}^j) = \beta \exp\left(-\frac{1}{2}\sum_k \lambda_k \left[(\mathbf{f}_k^i - \mathbf{f}_k^j)^2 + 1_{i \neq j}\sigma^2\right]\right).$$

This kernel is a variant of the radial basis function kernel, regularized by input noise $\sigma^2$ (since the outputs are noiseless). The GP covariance is then defined as $\mathbf{K}_{ij} = k(\mathbf{f}^i, \mathbf{f}^j)$, producing the following GP likelihood:

$$\log P(\mathbf{y}, \lambda, \beta|\mathbf{F}) = -\frac{1}{2}\mathbf{y}^{\mathrm{T}}\mathbf{K}^{-1}\mathbf{y} - \frac{1}{2}\log|\mathbf{K}| + \log P(\lambda, \beta|\mathbf{F}).$$

The last term in the likelihood is the prior on the hyperparameters. This prior encourages the feature weights $\lambda$ to be sparse, and prevents degeneracies that occur as $\mathbf{y} \to 0$. The latter is accomplished with a prior that encodes the belief that no two inducing points are deterministically dependent, as captured by their partial correlation:

$$\log P(\lambda, \beta|\mathbf{F}) = -\frac{1}{2}\mathrm{tr}\left(\mathbf{K}^{-2}\right) - \sum_k \log\left(\lambda_k + 1\right)$$

This prior is discussed in more detail in previous work (Levine et al., 2011). The reward at a feature point $\mathbf{f}(\mathbf{x}_t, \mathbf{u}_t)$ is given by the GP posterior mean, and can be augmented with a set of linear features $\mathbf{f}_\ell$:

$$r(\mathbf{x}_t, \mathbf{u}_t) = \mathbf{k}_t\alpha + \theta^{\mathrm{T}}\mathbf{f}_\ell(\mathbf{x}_t, \mathbf{u}_t),$$

where $\alpha = \mathbf{K}^{-1}\mathbf{y}$, and $\mathbf{k}_t$ is a row vector corresponding to the covariance between $\mathbf{f}(\mathbf{x}_t, \mathbf{u}_t)$ and each inducing point $\mathbf{f}^i$, given by $\mathbf{k}_{ti} = k(\mathbf{f}(\mathbf{x}_t, \mathbf{u}_t), \mathbf{f}^i)$. The exact log likelihood, before the proposed approximation, is obtained by using the GP likelihood as a prior on the IOC likelihood in Equation 1:

$$\log P(\mathbf{u}|\mathbf{x}_0) = \sum_t r(\mathbf{x}_t, \mathbf{u}_t) - \log Z + \log P(\mathbf{y}, \lambda, \beta|\mathbf{F}).$$

Only the IOC likelihood is altered by the proposed approximation. The gradient and Hessian of the reward with respect to the states are

$$\hat{\mathbf{g}}_t = \frac{\partial \mathbf{k}_t}{\partial \mathbf{x}_t}\alpha + \hat{\mathbf{g}}_t^{(\ell)} \quad \text{and} \quad \hat{\mathbf{H}}_t = \frac{\partial^2 \mathbf{k}_t}{\partial \mathbf{x}_t^2}\alpha + \hat{\mathbf{H}}_t^{(\ell)},$$

and the kernel derivatives are given by

$$\frac{\partial \mathbf{k}_t}{\partial \mathbf{x}_t} = \sum_k \frac{\partial \mathbf{k}_t}{\partial \mathbf{f}_k^t}\hat{\mathbf{g}}_t^{(k)}$$

$$\frac{\partial^2 \mathbf{k}_t}{\partial \mathbf{x}_t^2} = \sum_k \frac{\partial \mathbf{k}_t}{\partial \mathbf{f}_k}\hat{\mathbf{H}}_t^{(k)} + \sum_{k_1, k_2}\frac{\partial^2 \mathbf{k}_t}{\partial \mathbf{f}_{k_1}^t \partial \mathbf{f}_{k_2}^t}\hat{\mathbf{g}}_t^{(k_1)}\hat{\mathbf{g}}_t^{(k_2)}.$$

The feature derivatives $\hat{\mathbf{g}}^{(k)}$ and $\hat{\mathbf{H}}^{(k)}$ are defined in the previous section, and $\hat{\mathbf{g}}$ and $\hat{\mathbf{H}}$ are given analogously. Using these quantities, the likelihood can be computed as described in Section 5. The likelihood gradients are derived in Appendix C of the supplement.

This algorithm can learn more expressive rewards in domains where a linear reward basis is not known, but with the usual bias and variance tradeoff that comes with increased model complexity. As shown in our evaluation, the linear method requires fewer examples when a linear basis is available, while the nonlinear variant can work with much less expressive features.

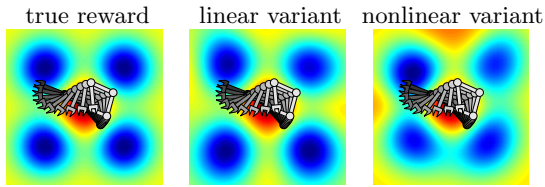

true reward    linear variant    nonlinear variant

*Figure 2.* Robot arm rewards learned by our algorithms for a 4-link arm. One of eight examples is shown.

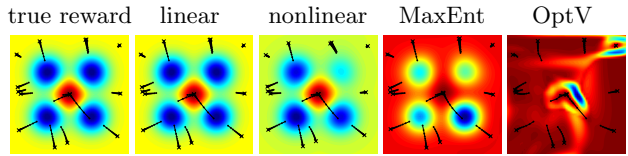

true reward   linear   nonlinear   MaxEnt   OptV

*Figure 3.* Planar navigation rewards learned from 16 locally optimal examples. Black lines show optimal paths for each reward originating from example initial states. The rewards learned by our algorithms better resemble the true reward than those learned by prior methods.

## 7. Evaluation

We evaluate our method on simulated robot arm control, planar navigation, and simulated driving. In the robot arm task, the expert sets continuous torques on each joint of an $n$-link planar robot arm. The reward depends on the position of the end-effector. Each link has an angle and a velocity, producing a state space with $2n$ dimensions. By changing the number of links, we can vary the dimensionality of the task. An example of a 4-link arm is shown in Figure 2. The complexity of this task makes it difficult to compare with prior work, so we also include a simple planar navigation task, in which the expert takes continuous steps on a plane, as shown in Figure 3. Finally, we use human-created examples on a simulated driving task, which shows how our method can learn complex policies from human demonstrations on a more realistic domain.

The reward function in the robot arm and navigation tasks has a Gaussian peak in the center, surrounded by four pits. The reward also penalizes each action with the square of its magnitude. The IOC algorithms are provided with a grid of 25 evenly spaced Gaussian features and the squared action magnitude. In the nonlinear test in Section 7.2, the Cartesian coordinates of the arm end-effector are provided instead of the grid.

We compare the linear and nonlinear variants of our method with the MaxEnt IRL and OptV algorithms (Ziebart et al., 2008; Dvijotham & Todorov, 2010). We present results for the linear time algorithm in Section 5.1, though we found that both the LQR variant and the direct, non-linearized approach produced similar results. MaxEnt used a grid discretization for both states and actions, while OptV used discretized actions and adapted the value function features as described by Dvijotham and Todorov. Since OptV cannot learn action-dependent rewards, it was provided with the true weight for the action penalty term.

To evaluate a learned reward, we first compute the optimal paths with respect to this reward from 32 random initial states that are not part of the training set. We also find the paths that begin in the same initial states but are optimal with respect to the true reward. In both cases, we compute evaluation paths that are globally optimal, by first solving a discretization of the task with value iteration, and then finetuning the paths with continuous optimization. Once the evaluation paths are computed for both reward functions, we obtain a reward loss by subtracting the true reward along the learned reward's path from the true reward along the true optimal path. This loss is low when the learned reward induces the same policy as the true one, and high when the learned reward causes costly mistakes. Since the reward loss is measured entirely on globally optimal paths, it captures how well each algorithm learns the true, global reward, regardless of whether the examples are locally or globally optimal.

### 7.1. Locally Optimal Examples

To test how well each method handles locally optimal examples, we ran the navigation task with increasing numbers of examples that were either globally or locally optimal. As discussed above, globally optimal examples were obtained with discretization, while locally optimal examples were computed by optimizing the actions from a random initialization. Each test was repeated eight times with random initial states for each example.

The results in Figure 4 show that both variants of our algorithm converge to the correct policy. The linear variant requires fewer examples, since the features form a good linear basis for the true reward. MaxEnt assumes global optimality and does not converge to the correct policy when the examples are only locally optimal. It also suffers from discretization error. OptV globally optimal examples     locally optimal examples has difficulty generalizing the reward to unseen parts of the state space, because the value function features do not impose meaningful structure on the reward.

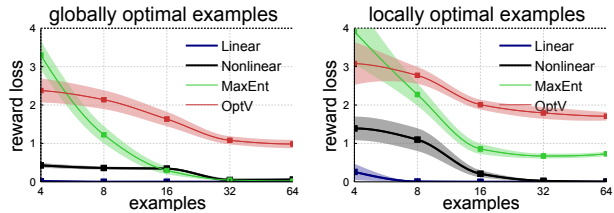

*Figure 4.* Reward loss for each algorithm with either globally or locally optimal planar navigation examples. Prior methods do not converge to the expert's policy when the examples are not globally optimal.

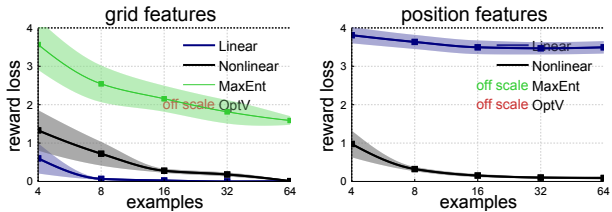

Figure 5. Reward loss for each algorithm with the Gaussian grid and end-effector position features on the 2-link robot arm task. Only the nonlinear variant of our method could learn the reward using only the position features.

| style | path | average speed | time behind | time in front |
|---|---|---|---|---|
| aggressive | learned | 158.1 kph | 3.5 sec | 12.5 sec |
| | human | 158.2 kph | 3.5 sec | 16.7 sec |
| evasive | learned | 150.1 kph | 7.2 sec | 3.7 sec |
| | human | 149.5 kph | 4.5 sec | 2.8 sec |
| tailgater | learned | 97.5 kph | 111.0 sec | 0.0 sec |
| | human | 115.3 kph | 99.5 sec | 7.0 sec |

Table 1. Statistics for sample paths for the learned driving rewards and the corresponding human demonstrations starting in the same initial states. The statistics of the learned paths closely resemble the holdout demonstrations.

### 7.2. Linear and Nonlinear Rewards

On the robot arm task, we evaluated each method with both the Gaussian grid features, and simple features that only provide the position of the end effector, and therefore do not form a linear basis for the true reward. The examples were globally optimal. The number of links was set to 2, resulting in a 4-dimensional state space. Only the nonlinear variant of our algorithm could successfully learn the reward from the simple features, as shown in Figure 5. Even with the grid features, which do form a linear basis for the reward, MaxEnt suffered greater discretization error due to the complex dynamics of this task, while OptV could not meaningfully generalize the reward due to the increased dimensionality of the task.

### 7.3. High Dimensional Tasks

To evaluate the effect of dimensionality, we increased the number of robot arm links. As shown in Figure 6, the processing time of our methods scaled gracefully with the dimensionality of the task, while the quality of the reward did not deteriorate appreciably. The processing time of OptV increased exponentially due to the action space discretization. The MaxEnt discretization was intractable with more than two links, and is therefore not shown.

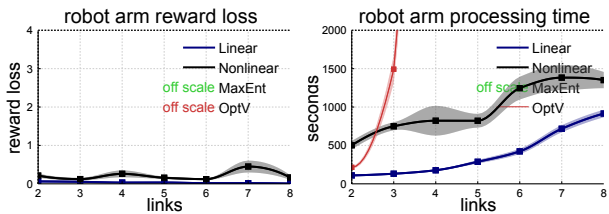

Figure 6. Reward loss and processing time with increasing numbers of robot arm links $n$, corresponding to state spaces with $2n$ dimensions. Our methods efficiently learn good rewards even as the dimensionality is increased.

### 7.4. Human Demonstrations

We evaluate how our method handles human demonstrations on a simulated driving task. Although driving policies have been learned by prior IOC methods (Abbeel & Ng, 2004; Levine et al., 2011), their discrete formulation required a discrete simulator where the agent makes simple decisions, such as choosing which lane to switch to. In constrast, our driving simulator is a fully continuous second order dynamical system. The actions correspond directly to the gas, breaks, and steering of the simulated car, and the state space includes position, orientation, and linear and angular velocities. Because of this, prior methods that rely on discretization cannot tractably handle this domain.

We used our nonlinear method to learn from sixteen 13-second examples of an aggressive driver that cuts off other cars, an evasive driver that drives fast but keeps plenty of clearance, and a tailgater who follows closely behind the other cars. The features are speed, deviation from lane centers, and Gaussians covering the front, back, and sides of the other cars on the road. Since there is no ground truth reward for these tasks, we cannot use the reward loss metric. We follow prior work and quantify how much the learned policy resembles the demonstration by using task-relevant statistics (Abbeel & Ng, 2004). We measure the average speed of sample paths for the learned reward and the amount of time they spend within two car-lengths behind and in front of other cars, and compare these statistics with those from an unobserved holdout set of user demonstrations that start in the same initial states. The results in Table 1 show that the statistics of the learned policies are similar to the holdout demonstrations.

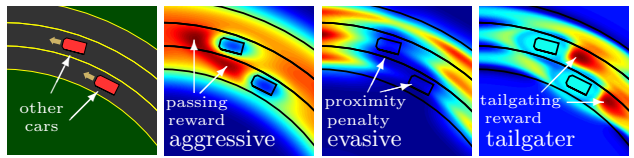

Figure 7. Learned driving style rewards. The aggressive reward is high in front of other cars, the evasive one avoids them, and the tailgater reward is high behind the cars.

Plots of the learned rewards are shown in Figure 7. Videos of the optimal paths for the learned rewards can be downloaded from the project website, along with the supplementary appendices and source code: http://graphics.stanford.edu/projects/cioc.

## 8. Discussion and Future Work

We presented an IOC algorithm designed for continuous, high dimensional domains. Our method remains efficient in high dimensional domains by using a local approximation to the reward function likelihood. This approximation also removes the global optimality requirement for the expert's demonstrations, allowing the method to learn the reward from examples that are only locally optimal. Local optimality can be easier to demonstrate than global optimality, particularly in high dimensional domains. As shown in our evaluation, prior methods do not converge to the underlying reward function when the examples are only locally optimal, regardless of how many examples are provided.

Since our algorithm relies on the derivatives of the reward features to learn the reward function, we require the features to be differentiable with respect to the states and actions. These derivatives must be precomputed only once, so it is quite practical to use finite differences when analytic derivatives are unavailable, but features that exhibit discontinuities are still poorly suited for our method. Our current formulation also only considers deterministic, fixed-horizon control problems, and an extension to stochastic or infinite-horizon cases is an interesting avenue for future work.

In addition, although our method handles examples that lack global optimality, it does not make use of global optimality when it is present: the examples are always assumed to be only locally optimal. Prior methods that exploit global optimality can infer more information about the reward function from each example when the examples are indeed globally optimal.

An exciting avenue for future work is to apply this approach to other high dimensional, continuous problems that have previously been inaccessible for inverse optimal control methods. One challenge with such applications is to generalize and impose meaningful structure in high dimensional tasks without requiring detailed features or numerous examples. While the nonlinear variant of our algorithm takes one step in this direction, it still relies on features to generalize the reward to unseen regions of the state space. A more sophisticated way to construct meaningful, generalizable features would allow IOC to be easily applied to complex, high dimensional tasks.

## Acknowledgements

We thank the anonymous reviewers for their constructive comments. Sergey Levine was supported by NSF Graduate Research Fellowship DGE-0645962.

## References

Abbeel, Pieter and Ng, Andrew Y. Apprenticeship learning via inverse reinforcement learning. In *Proceedings of ICML*, 2004.

Birgin, Ernesto G. and Martínez, José Mario. Practical augmented lagrangian methods. In *Encyclopedia of Optimization*, pp. 3013–3023. 2009.

Boyd, S., El Ghaoui, L., Feron, E., and Balakrishnan, V. *Linear Matrix Inequalities in System and Control Theory*. SIAM, Philadelphia, PA, June 1994.

Dvijotham, Krishnamurthy and Todorov, Emanuel. Inverse optimal control with linearly-solvable MDPs. In *Proceedings of ICML*, 2010.

Levine, Sergey, Popović, Zoran, and Koltun, Vladlen. Feature construction for inverse reinforcement learning. In *Advances in Neural Information Processing Systems*. 2010.

Levine, Sergey, Popović, Zoran, and Koltun, Vladlen. Nonlinear inverse reinforcement learning with gaussian processes. In *Advances in Neural Information Processing Systems*. 2011.

Ng, Andrew Y. and Russell, Stuart J. Algorithms for inverse reinforcement learning. In *Proceedings of ICML*, 2000.

Ratliff, Nathan, Bagnell, J. Andrew, and Zinkevich, Martin A. Maximum margin planning. In *Proceedings of ICML*, 2006.

Ratliff, Nathan, Silver, David, and Bagnell, J. Andrew. Learning to search: Functional gradient techniques for imitation learning. *Autonomous Robots*, 27(1): 25–53, 2009.

Tierney, Luke and Kadane, Joseph B. Accurate approximations for posterior moments and marginal densities. *Journal of the American Statistical Association*, 81(393):82–86, 1986.

Ziebart, Brian D. *Modeling Purposeful Adaptive Behavior with the Principle of Maximum Causal Entropy*. PhD thesis, Carnegie Mellon University, 2010.

Ziebart, Brian D., Maas, Andrew, Bagnell, J. Andrew, and Dey, Anind K. Maximum entropy inverse reinforcement learning. In *AAAI Conference on Artificial Intelligence*, 2008.